# Existential Rule Languages with Finite Chase: Complexity and Expressiveness

**Heng Zhang**[1] and **Yan Zhang**[1] and **Jia-Huai You**[2]

[1]School of Computing, Engineering and Mathematics, University of Western Sydney, Penrith, NSW 2751, Australia
[2]Department of Computing Science, University of Alberta, Edmonton, AB T6G 2E8, Canada

## Abstract

Finite chase, or alternatively chase termination, is an important condition to ensure the decidability of existential rule languages. In the past few years, a number of rule languages with finite chase have been studied. In this work, we propose a novel approach for classifying the rule languages with finite chase. Using this approach, a family of decidable rule languages, which extend the existing languages with the finite chase property, are naturally defined. We then study the complexity of these languages. Although all of them are tractable for data complexity, we show that their combined complexity can be arbitrarily high. Furthermore, we prove that all the rule languages with finite chase that extend the weakly acyclic language are of the same expressiveness as the weakly acyclic one, while rule languages with higher combined complexity are in general more succinct than those with lower combined complexity.

## Introduction

It has been shown that existential rule languages, also called Datalog$^{\pm}$, have prominent applications in ontological reasoning, knowledge representation, and databases, in which query answering is a primary yet challenging problem; see e.g., (Calì et al. 2010; Baget et al. 2011a). Under an existential rule language, queries are answered against a logical theory consisting of an input database and a finite set of existential rules, while a chase procedure is usually used. Specifically, given an input database $D$, a finite set $\Sigma$ of existential rules, and a query $q$, we want to decide whether $D \cup \Sigma \models q$. Applying the chase procedure, the problem is equivalent to deciding whether $\mathsf{chase}(D, \Sigma) \models q$. Through a chase procedure, fresh *nulls* have to be introduced for each application of the existential rules, and hence, potential cyclic applications of these rules may lead the chase procedure not to terminate, i.e., $\mathsf{chase}(D, \Sigma)$ is infinite. Therefore, the problem of query answering under existential rule languages is in general undecidable (Beeri and Vardi 1981).

There have been a considerable number of works on identifying decidable classes with respect to query answering. Basically, two major approaches have provided a landscape on this study: One is to focus on some restricted fragments of existential rule languages such that the underlying chase procedure, though non-terminating in general, still enjoys some kind of *finite representability property*, so that the problem of query answering is decidable under this setting. This paradigm includes, e.g., *guarded rules* (Calì, Gottlob, and Lukasiewicz 2012), *greedy bounded treewidth sets* (Baget et al. 2011b), *sticky sets* (Calì, Gottlob, and Pieris 2012), and *Shy programs* (Leone et al. 2012). The other approach is to identify a certain *acyclicity* condition under which each existential rule can only be finitely applied so that the chase procedure always terminates. There have been many recent studies on this paradigm. Our work presented in this paper is along this line. Below, let us provide a brief summary of recent works under this approach.

In their milestone paper, Fagin et al. (2005) formulated a concept called *weak acyclicity* (WA) as a sufficient condition to ensure the chase termination for existential rules. This concept was then extended to a number of notions, such as *stratification* (Deutsch, Nash, and Remmel 2008), *super-weak acyclicity* (Marnette 2009), *local stratification* (Greco, Spezzano, and Trubitsyna 2011), *joint acyclicity* (Krötzsch and Rudolph 2011), *model-faithful acyclicity* (MFA) and *model-summarising acyclicity* (MSA) (Grau et al. 2013), and some dependency relations by (Baget et al. 2014). Among these, MFA is known to define the largest rule class. In addition, many ontologies in various domains turn out to be in the MFA class, as evidenced in (Grau et al. 2013).

It has been observed that almost all of the existential rule languages defined based on the notion of acyclicity or its variations have PTIME-complete data complexity and 2-EXPTIME-complete combined complexity. The uniformity on data complexity is in fact due to an interesting result proved in (Marnette 2009), which states that every rule language with finite Skolem chase is in PTIME for data complexity. A natural question then arises: Does this uniformity hold for combined complexity? Moreover, what is the expressiveness of existing rule languages with finite chase? Please note that the uniformity on data complexity does not imply the uniformity on expressiveness as data complexity only captures the hardest case of a language. Recently, there have been two interesting related works that studied the expressiveness of existential rules (Arenas, Gottlob, and Pieris 2014; Gottlob, Rudolph, and Simkus 2014). But both of them only focus on guarded language or its variations.

In this paper, we study the complexity, expressiveness and succinctness for existential rule languages with finite chase. Our contributions are summarized as follows:

1. A novel approach for classifying the existential rule languages with finite Skolem chase is proposed by restricting the use of existential variables in the Skolem chase. Under this approach, a family of interesting decidable rule languages, called *bounded languages*, are naturally defined. All of the existing rule languages with finite chase, e.g., the MFA class, are contained in these languages.

2. For every nonnegative integer $k$, the combined complexity of Boolean query answering for $k$-exponentially bounded language is proved to be $(k+2)$-EXPTIME-complete, and the membership problem of $k$-exponentially bounded language is proved to be in $(k+2)$-EXPTIME. Furthermore, for other bounded languages, the corresponding upper bounds of the complexity are also obtained.

3. All the languages with finite Skolem chase that extend the WA class are proved to be of the same expressiveness as WA, while languages with higher combined complexity are in general more succinct than those with lower combined complexity. On ordered databases, WA is shown to capture all existential rule sets whose universal models are computable in PTIME, even if they have no finite chase.

The results presented in this paper not only generalize some of the existing works, such as the two acyclicity notions of MFA and MSA proposed in (Grau et al. 2013), more importantly, they provide a global landscape for characterizing the existential rule languages with finite Skolem chase.

The rest of this paper is organized as follows. Section 2 provides necessary preliminaries. Section 3 defines a family of existential rule languages with finite Skolem chase called *bounded classes*, and presents some interesting properties of this family of languages. Section 4 then focuses on the complexity issues for bounded classes of languages, while section 5 explores the expressiveness and succinctness of these bounded classes of languages in details. Finally, section 6 concludes this paper with some remarks. Due to space limitation, proofs of some results are presented in an extended version of this paper, see (Zhang, Zhang, and You 2014).

## Preliminaries

**Databases and Queries.** As usual, we assume (i) an infinite set $\Delta$ of *constants*, (ii) an infinite set $\Delta_n$ of *(labelled) nulls*, and (iii) an infinite set $\Delta_v$ of *variables*. A *relational schema* $\mathcal{R}$ consists of a finite set of *relation symbols*, each of which is armed with a natural number, its *arity*. *Terms* are either constants or variables. Every *atomic formula* (or *atom*) has the form $R(t)$ where $R$ is a relation symbol and $t$ a tuple of terms of a proper length. *Ground terms* are terms involving no variable, and *facts* are atoms built from ground terms.

Given a relational schema $\mathcal{R}$, an *instance (database)* over $\mathcal{R}$, or simply $\mathcal{R}$-*instance* ($\mathcal{R}$-*database*), is a (finite) set of facts involving only relation symbols from $\mathcal{R}$. The *domain* of a database $D$, denoted $\text{dom}(D)$, is the set of all constants appearing in $D$. *General instances (databases)* are the extensions of instances (databases) by allowing nulls to be used. Given a general instance (database) $D$ and a relational schema $\mathcal{R}$, the *restriction* of $D$ to $\mathcal{R}$, denoted $D|_{\mathcal{R}}$, is the set of facts in $D$ involving only relation symbols from $\mathcal{R}$.

A *substitution* is a function $h : \Delta \cup \Delta_n \cup \Delta_v \to \Delta \cup \Delta_n \cup \Delta_v$ with (i) $h(c) = c$ for all $c \in \Delta$ and (ii) $h(\mathsf{n}) \in \Delta \cup \Delta_n$ for all $\mathsf{n} \in \Delta_n$. Let $D$ and $D_0$ be general instances of the same schema. Then $D$ is called *homomorphic* to $D_0$, written $D \to D_0$, if there is a substitution $h$ with $h(D) \subseteq D_0$ where $h$ is assumed to be extended to atoms and general instances naturally. In this case, the function is called a *homomorphism* from $D$ to $D_0$. Moreover, $D$ is *homomorphically equivalent* to $D_0$ if $D$ is homomorphic to $D_0$ and vice versa.

Every *conjunctive query (CQ)* $q$ over a relational schema $\mathcal{R}$ has the form $q(\boldsymbol{x}) := \exists \boldsymbol{y} \varphi(\boldsymbol{x}, \boldsymbol{y})$, where $\boldsymbol{x}, \boldsymbol{y}$ are tuples of variables, and $\varphi(\boldsymbol{x}, \boldsymbol{y})$ is a conjunction (sometimes we regard it as a set) of atoms with variables from $\boldsymbol{x}$ and $\boldsymbol{y}$, and relation symbols from $\mathcal{R}$. A *Boolean CQ (BCQ)* is a CQ of the form $q()$. Actually, BCQs can be regarded as general databases if we omit the quantifiers and regard the variables as nulls. Given any BCQ $q$ and any general instance $D$ over the same schema, the *answer* to $q$ over $D$ is "*Yes*", written $D \models q$, if there exists a homomorphism from $q$ to $D$.

**Existential Rules and Skolem Chase.** Given a relational schema $\mathcal{R}$, every *(existential) rule* over $\mathcal{R}$ is a first-order sentence $\gamma$ of the form $\forall \boldsymbol{x} \forall \boldsymbol{y}(\varphi(\boldsymbol{x}, \boldsymbol{y}) \to \exists \boldsymbol{z} \psi(\boldsymbol{x}, \boldsymbol{z}))$, where $\varphi$ and $\psi$ are conjunctions of atoms with relation symbols from $\mathcal{R}$ and variables from $\boldsymbol{x} \cup \boldsymbol{y}$ and $\boldsymbol{x} \cup \boldsymbol{z}$, respectively. We call $\varphi$ the *body* of $\gamma$ and $\psi$ the *head* of $\gamma$, and write them as $\text{body}(\gamma)$ and $\text{head}(\gamma)$, respectively. When writing a rule, for simplicity, we will omit the universal quantifiers.

A *rule ontology* is a triple $(\Sigma, \mathcal{D}, \mathcal{Q})$, where $\Sigma$ is *finite* and *nonempty* set $\Sigma$ of rules, $\mathcal{D}$, called *database schema*, is a relational schema consisting of the relation symbols to be used in databases, and $\mathcal{Q}$, called *query schema*, is a relational schema consisting of the relation symbols to be used in queries. Relation symbols appearing in $\Sigma$ but neither $\mathcal{D}$ nor $\mathcal{Q}$ are called *auxiliary symbols*. Note that $\mathcal{D}$ and $\mathcal{Q}$ could be the same. Without loss of generality, in any rule ontology, each variable is assumed to be quantified at most once.

Let $\gamma$ be a rule $\varphi(\boldsymbol{x}, \boldsymbol{y}) \to \exists \boldsymbol{z} \psi(\boldsymbol{x}, \boldsymbol{z})$. We introduce a function symbol $\mathsf{f}_z$ of arity $|\boldsymbol{x}|$ for each variable $z \in \boldsymbol{z}$. From now on, we will regard terms built from constants and the introduced function symbols as a special class of nulls. The *functional transformation* of $\gamma$, denoted $\mathsf{sk}(\gamma)$, is the formula obtained from $\gamma$ by substituting $\mathsf{f}_z(\boldsymbol{x})$ for each variable $z \in \boldsymbol{z}$. Given a set $\Sigma$ of rules, the *functional transformation* of $\Sigma$, denoted $\mathsf{sk}(\Sigma)$, is the set of rules $\mathsf{sk}(\gamma)$ for all $\gamma \in \Sigma$.

Now we are in the position to define the *(Skolem) chase*. Let $D$ be a database and $\Sigma$ a rule set. We let $\text{chase}^0(D, \Sigma) = D$ and, for all $n > 0$, let $\text{chase}^n(D, \Sigma)$ denote the union of $\text{chase}^{n-1}(D, \Sigma)$ and $h(\text{head}(\gamma))$ for all rules $\gamma \in \mathsf{sk}(\Sigma)$ and all substitutions $h$ such that $h(\text{body}(\gamma)) \subseteq \text{chase}^{n-1}(D, \Sigma)$. Let $\text{chase}(D, \Sigma)$ be the union of $\text{chase}^n(D, \Sigma)$ for all $n \geq 0$. It is well-known that, for all BCQs $q$, $D \cup \Sigma \models q$ (under the semantics of first-order logic) if and only if $\text{chase}(D, \Sigma) \models q$. Given a rule ontology $O = (\Sigma, \mathcal{D}, \mathcal{Q})$, we say that $O$ has *finite chase* if for all $\mathcal{D}$-databases $D$, $\text{chase}(D, \Sigma)$ is finite. For more details, please refer to (Marnette 2009).

**More Notations.** Given a set $\Sigma$ of rules and a BCQ $q$, let $\|\Sigma\|$ and $\|q\|$ denote the numbers of symbols occurring in $\Sigma$ and $q$, respectively. We assume that the reader is familiar with complexity theory. Given a unary function $T$ on natural numbers, by $\text{DTIME}(T(n))$ we mean the class of complexity languages decidable in time $T(n)$ by a deterministic Turing machine. For $k \geq 0$ we let $\exp_k(n)$ denote the function that maps $n$ to $n$ if $k = 0$, and $2^{\exp_{k-1}(n)}$ otherwise. By $k$-EXPTIME we mean the class $\bigcup_{c \geq 0} \text{DTIME}(\exp_k(n^c))$.

For simplicity, we denote relation symbols (nulls/function symbols, respectively) by capitalized (lower-case, respectively) sans-serif letters, constants by lower-case italic letters $a, b, c$, variables by lower-case italic letters $u, v, w, x, y, z$, and terms by lower-case italic letters $s, t$. All of these symbols may be written with subscripts or superscripts. In addition, bold italic letters $\boldsymbol{u}, \boldsymbol{v}, \boldsymbol{w}, \boldsymbol{x}, \boldsymbol{y}, \boldsymbol{z}$ and $\boldsymbol{s}, \boldsymbol{t}$ are used to range over tuples of variables and terms, respectively.

## Bounded Classes

In this section, we define a family of rule languages with finite chase, and study its general properties.

We first define some notations. Given a ground term $t$, the *height* of $t$, denoted $\text{ht}(t)$, is defined as follows:

$$\text{ht}(t) := \begin{cases} 0 & \text{if } t \in \Delta; \\ \max\{\text{ht}(s) : s \in \boldsymbol{s}\} + 1 & \text{if } t = \mathsf{f}(\boldsymbol{s}) \text{ for some } \mathsf{f}. \end{cases}$$

Given any general instance $A$, the *height* of $A$, denoted $\text{ht}(A)$, is defined as the maximum height of terms that have at least one occurrence in $A$ if it exists, and $\infty$ otherwise.

**Definition 1.** Every *bound function* is a function from positive integers to positive integers. Let $\delta$ be a bound function. A rule ontology $(\Sigma, \mathcal{D}, \mathcal{Q})$ is called $\delta$-*bounded* if, for all $\mathcal{D}$-databases $D$, $\text{ht}(\text{chase}(D, \Sigma)) \leq \delta(\|\Sigma\|)$. We let $\delta$-BOUNDED denote the class of $\delta$-bounded rule ontologies.

As there exist an infinite number of bound functions, it is interesting to know if there is a "maximum" bound function that captures all $\delta$-bounded rule ontologies for any bound function $\delta$ (or all rule ontologies with finite chase). The following result shows that the answer is definitely "yes".

**Proposition 1.** *There is a bound function $\delta$ such that, for every rule ontology $O$, $O$ has finite chase iff it is $\delta$-bounded.*

*Proof.* (Sketch) We first construct a bound function $\delta$, and then it suffices to show that every rule ontology with finite chase is $\delta$-bounded. To define $\delta$, we want to prove that, for every rule ontology $O = (\Sigma, \mathcal{D}, \mathcal{Q})$, there exists a database $D_O^*$ such that $\text{ht}(\text{chase}(D, \Sigma)) \leq \text{ht}(\text{chase}(D_O^*, \Sigma))$ for all $\mathcal{D}$-databases $D$. This can be done by employing the so-called *critical database technique*, which was developed in (Marnette 2009). Define $\delta(n)$ as the maximum $\text{ht}(\text{chase}(D_O^*, \Sigma))$ among all rule ontologies $O = (\Sigma, \mathcal{D}, \mathcal{Q})$ with finite chase such that $\|\Sigma\| \leq n$; we then have the desired function $\delta$. $\square$

*Remark* 1. Let BOUNDED be the union of $\delta$-BOUNDED for all bound functions $\delta$. A rule ontology is called *bounded* if it belongs to BOUNDED. As all bounded rule ontologies have finite chase, by Proposition 1 we have that BOUNDED contains exactly the rule ontologies with finite chase.

Next, let us define a class of interesting bound functions.

**Definition 2.** Let $k$ be a natural number and let $\exp_k$ be the function defined in the previous section. A rule ontology is called $k$-*exponentially bounded* if it is $\exp_k$-bounded.

*Remark* 2. The MFA class (Grau et al. 2013), which was shown to extend many existing languages with an acyclicity restriction, is defined by restricting the recursive uses of existential variables in Skolem chase. It is not difficult to see that MFA $\subseteq \exp_0$-BOUNDED. The following example shows that the inclusion is in fact strict.

**Example 1.** Let $O = (\Sigma, \mathcal{D}, \mathcal{Q})$ be a rule ontology, where $\mathcal{D} = \{\mathsf{R}\}$ and $\Sigma$ consists of the following rules:

$$\mathsf{R}(x, x) \rightarrow \exists yz\, \mathsf{S}(x, y) \wedge \mathsf{S}(y, z)$$
$$\mathsf{R}(x, y) \wedge \mathsf{S}(x, z) \rightarrow \exists v\, \mathsf{R}(z, v)$$

This rule ontology does not belong to the MFA class because the existential variable $v$ might be recursively applied in the Skolem chase (one can check it by letting the database $D$ be $\{\mathsf{R}(a, a)\}$). As each existential variable can be recursively used at most twice, $O$ is 0-exponentially bounded. $\square$

One might ask if all bounded ontologies can be captured by exponential bound functions (or computable bound functions). The proposition below shows that this is impossible.

**Proposition 2.** *There is no computable bound function $\delta$ such that every bounded rule ontology is $\delta$-bounded.*

## Complexity

Now we study the complexity of bounded classes. We are interested in the complexity of two kinds of important computations: query answering and language membership.

**Boolean Query Answering.** The problem to be investigated here, also known as *query entailment*, is defined as follows: Given a set $\Sigma$ of rules, a database $D$ and a Boolean query $q$, decide if $D \cup \Sigma \models q$. We first consider the upper bound.

**Proposition 3.** *Let $\delta$ be a bound function. Then for any $\delta$-bounded rule ontology $(\Sigma, \mathcal{D}, \mathcal{Q})$, any $\mathcal{D}$-database $D$ and any BCQ $q$ over $\mathcal{Q}$, it is in*

$$\text{DTIME}((|\text{dom}(D)| + \|\Sigma\|)^{\|\Sigma\|^{\mathcal{O}(\delta(\|\Sigma\|))} \cdot \|q\|^{\mathcal{O}(1)}})$$

*to check whether $D \cup \Sigma \models q$.*

*Proof.* (Sketch) First evaluate the size of $\text{chase}(D, \Sigma)$. By this we know how many stages are needed for the chase to terminate. Counting the cost of each chase stage and querying on $\text{chase}(D, \Sigma)$, we then have the desired result. $\square$

A lower bound for the combined complexity is as follows.

**Proposition 4.** *It is $(k + 2)$-EXPTIME-hard (for the combined complexity) to check, given a $k$-exponentially bounded rule ontology $(\Sigma, \mathcal{D}, \mathcal{Q})$, a $\mathcal{D}$-database $D$ and a BCQ $q$ over $\mathcal{Q}$, whether $D \cup \Sigma \models q$.*

*Proof.* (Sketch) Let $M$ be any deterministic Turing machine that terminates in $\exp_{k+2}(n)$ steps on any input of length $n$. Let $\mathcal{D} = \emptyset$ and $\mathcal{Q} = \{\mathsf{Accept}\}$ where $\mathsf{Accept}$ is a nullary relation symbol. To show the desired result, it suffices to show that, for each input (a binary string) $x$, there is an $\exp_k$-bounded rule ontology $(\Sigma, \mathcal{D}, \mathcal{Q})$ such that $M$ terminates on input $x$ if and only if $\emptyset \cup \Sigma \models \mathsf{Accept}$. Let $x$ be an input of length $n$. To construct the rule set $\Sigma$, the main difficulty is to define a linear order of length $\exp_{k+2}(n)$. If the order is defined, by an encoding similar to that in (Dantsin et al. 2001), one can construct a set of datalog rules to encode both $M$ and $x$. Here we only explain how to define the linear order.

Let us first consider the case where $k$ is even. The general idea is to construct a sequence of rule sets $(\Sigma_i)_{i \geq 0}$. For each $i$, let $\mathsf{Succ}_i, \mathsf{Min}_i$ and $\mathsf{Max}_i$ be relation symbols intended to define the (immediate) successor relation, the minimum element and the maximum element, respectively, of a linear order. For $i > 0$, the function of $\Sigma_i$ is as follows: If $\mathsf{Succ}_{i-1}, \mathsf{Min}_{i-1}$ and $\mathsf{Max}_{i-1}$ define a linear order of length $n$, then $\mathsf{Succ}_i, \mathsf{Min}_i$ and $\mathsf{Max}_i$ define a linear order on integers (represented in binary strings) from 0 to $2^{2^n}$. To implement each $\Sigma_i$, we generalize a technique used in the proof of Theorem 1 in (Calì, Gottlob, and Pieris 2010).

The first task is to define the binary strings of length one, i.e. "0" and "1". This can be done by the following rule:
$$\mathsf{Min}_{i-1}(v) \to \mathsf{BS}_i(v, 0) \land \mathsf{BS}_i(v, 1)$$
where $\mathsf{BS}_i(v, x)$ states that $x$ is a binary string of length $2^v$.

The following rules are used to generate binary strings of length $2^{v+1}$ by combining two binary strings of length $2^v$:
$$\mathsf{BS}_i(v, x) \land \mathsf{BS}_i(v, y) \;\to\; \exists z \, \mathsf{C}_i(v, x, y, z)$$
$$\mathsf{C}_i(v, x, y, z) \land \mathsf{Succ}_{i-1}(v, w) \;\to\; \mathsf{BS}_i(w, z)$$

Then, some rules to define a successor relation (w.r.t. the lexicographic order) on strings of length $2^{v+1}$ are as follows:
$$\begin{bmatrix} \mathsf{C}_i(v, x, y, z) \land \mathsf{C}_i(v, x, y_0, z_0) \\ \land\, \mathsf{Succ}^*_i(v, y, y_0) \land \mathsf{Succ}_{i-1}(v, w) \end{bmatrix} \to \mathsf{Succ}^*_i(w, z, z_0)$$
$$\begin{bmatrix} \mathsf{C}_i(v, x, y, z) \land \mathsf{C}_i(v, x_0, y_0, z_0) \\ \land\, \mathsf{Max}^*_i(v, y) \land \mathsf{Min}^*_i(v, y_0) \\ \land\, \mathsf{Succ}^*_i(v, x, x_0) \land \mathsf{Succ}_{i-1}(v, w) \end{bmatrix} \to \mathsf{Succ}^*_i(w, z, z_0)$$
where $\mathsf{Succ}^*_i(v, x, y)$ is intended to assert that $y$ is the immediate successor of $x$, and both $x$ and $y$ are of length $2^v$.

The minimum and the maximum binary strings of length $2^{v+1}$ are defined by the following rules:
$$\begin{bmatrix} \mathsf{Min}^*_i(v, x) \land \mathsf{Min}^*_i(v, y) \\ \land\, \mathsf{C}_i(v, x, y, z) \land \mathsf{Succ}_{i-1}(v, w) \end{bmatrix} \to \mathsf{Min}^*_i(w, z)$$
$$\begin{bmatrix} \mathsf{Max}^*_i(v, x) \land \mathsf{Max}^*_i(v, y) \\ \land\, \mathsf{C}_i(v, x, y, z) \land \mathsf{Succ}_{i-1}(v, w) \end{bmatrix} \to \mathsf{Max}^*_i(w, z)$$

Now the desired relations $\mathsf{Num}_i, \mathsf{Succ}_i, \mathsf{Min}_i$ and $\mathsf{Max}_i$ can be obtained by applying the following rules:
$$\mathsf{Succ}^*_i(v, x, y) \land \mathsf{Max}_{i-1}(v) \;\to\; \mathsf{Succ}_i(x, y)$$
$$\mathsf{Min}^*_i(v, x) \land \mathsf{Max}_{i-1}(v) \;\to\; \mathsf{Min}_i(x)$$
$$\mathsf{Max}^*_i(v, x) \land \mathsf{Max}_{i-1}(v) \;\to\; \mathsf{Max}_i(x)$$

For all $i > 0$, let $\Sigma_i$ consist of all of the above rules. It is easy to see that $\Sigma_i$ is as desired. Let $0, \ldots, n-1$ be distinct constants. Let $\Sigma_0$ denote the following rule set:
$$\mathsf{Min}_0(0) \land \mathsf{Max}_0(n-1)$$
$$\mathsf{Succ}_0(0, 1) \land \cdots \land \mathsf{Succ}_0(n-2, n-1)$$

Next, let $\ell = k/2 + 1$ and let $\Sigma_{\mathsf{num}}$ be the union of $\Sigma_i$ for all $i$ with $0 \leq i \leq \ell$. By the previous analysis, it is not difficult to see that $\mathsf{Succ}_\ell, \mathsf{Min}_\ell$ and $\mathsf{Max}_\ell$ define a linear order on (the binary representations of) integers from 0 to $\exp_{k+2}(n)$. It is also not difficult to check that the rule ontology $(\Sigma_{\mathsf{num}}, \emptyset, \{\mathsf{Accept}\})$ is $\exp_k$-bounded.

For the case where $k$ is odd, we can achieve the goal by some slight modifications to $\Sigma_{\mathsf{num}}$: (i) substituting the least integer greater than or equal to $\log n$ for $n$ in $\Sigma_0$, and then (ii) setting $\ell = k/2 + 2$. Similarly, we can show that the resulting rule set $\Sigma_{\mathsf{num}}$ satisfies the desired property. □

Now, by combining Propositions 3, 4, and the data complexity of Datalog (see, e.g., (Dantsin et al. 2001)), for any $k$-exponentially bounded class we then have the exact bound of the complexity w.r.t. Boolean query answering.

**Theorem 5.** *For all integers $k \geq 0$, the Boolean query answering problem of the $k$-exponentially bounded language is $(k+2)$-EXPTIME-complete for the combined complexity, and PTIME-complete for the data complexity.*

**Membership of Language.** Now we consider the membership problem of bounded languages. The problem is as follows: Given a rule ontology, check whether it belongs to the bounded language under consideration. Since the boundedness is defined in a semantical way, it is interesting to know how to check whether a rule ontology is $\delta$-bounded.

**Proposition 6.** *Let $\delta$ be a bound function that is computable in $\mathrm{DTIME}(T(n))$ for some function $T(n)$. Then for every rule ontology $O = (\Sigma, \mathcal{D}, \mathcal{Q})$, it is in*
$$\mathrm{DTIME}(\|\Sigma\|^{\|\Sigma\|^{\mathcal{O}(\delta(\|\Sigma\|))}} + T(\log \|\Sigma\|)^{\mathcal{O}(1)})$$
*to check whether $O$ is $\delta$-bounded.*

The above proposition can be proved by using Marnette's critical database technique (2009) and then by an analysis similar to that in the proof of Proposition 3.

*Remark* 3. Two immediate corollaries of Proposition 6 are: It is in $(k+2)$-EXPTIME to check whether a rule ontology is $k$-exponentially bounded; moreover, the membership for $\delta$-bounded language is decidable whenever $\delta$ is computable.

## Expressiveness and Succinctness

Though all rule languages with finite chase are tractable for data complexity (Marnette 2009), in the last section we have shown that their combined complexity could be very high. Hence, a natural question is as follows: Are the languages with high combined complexity really necessary for representing ontological knowledge? In this section, we address this question on two aspects: What is the expressiveness of these languages? How about the succinctness among them?

**Universal Worldview Mapping.** We first propose a semantic (and more general) definition for rule ontologies.

**Definition 3.** Let $\mathcal{D}$ and $\mathcal{Q}$ be two relational schemas. A *universal worldview mapping*, or UWM for short, over $(\mathcal{D}, \mathcal{Q})$ is a function that maps every $\mathcal{D}$-database $D$ to a general instance $Q$ over $\mathcal{Q}$. Let $\Phi$ and $\Psi$ be two UWMs over $(\mathcal{D}, \mathcal{Q})$. We say that $\Phi$ is *equivalent* to $\Psi$, written $\Phi \approx \Psi$, if for all $\mathcal{D}$-databases, $\Phi(D)$ is homomorphically equivalent to $\Psi(D)$.

It is clear that $\approx$ is an equivalence relation on the UWMs. Next, we show how to define UWMs from rule ontologies.

**Definition 4.** Let $O = (\Sigma, \mathcal{D}, \mathcal{Q})$ be any rule ontology. We define $[\![O]\!]$ as the function that maps every $\mathcal{D}$-database $D$ to the general instance $\mathsf{chase}(D, \Sigma)|_\mathcal{Q}$.

Given any rule ontology $O$, it is clear that $[\![O]\!]$ is a UWM.

We say that two rule ontologies $O_1$ and $O_2$ are *equivalent* if the corresponding UWMs are equivalent, i.e., $[\![O_1]\!] \approx [\![O_2]\!]$. The following property explains why this is desired.

**Proposition 7.** *Let $O_1 = (\Sigma_1, \mathcal{D}, \mathcal{Q})$ and $O_2 = (\Sigma_2, \mathcal{D}, \mathcal{Q})$ be two rule ontologies with finite chase. Then $[\![O_1]\!] \approx [\![O_2]\!]$ iff, for all $\mathcal{D}$-databases $D$ and all BCQs $q$ over $\mathcal{Q}$, we have*

$$D \cup \Sigma_1 \models q \quad \textit{iff} \quad D \cup \Sigma_2 \models q.$$

In addition, for a technical reason, given a rule ontology $O = (\Sigma, \mathcal{D}, \mathcal{Q})$, we require that $\mathcal{D}$ and $\mathcal{Q}$ are *disjoint* and *no relation symbol in $\mathcal{D}$ has an occurrence in the head of any rule in $\Sigma$.*[1] We call such rule ontologies *normal*. These assumptions do not change the expressiveness since, for every relation symbol $\mathsf{R} \in \mathcal{D} \cap \mathcal{Q}$, we can always replace $\mathsf{R}$ in $\mathcal{D}$ with a fresh relation symbol $\mathsf{R}'$ of the same arity, and then add a *copy rule* $\mathsf{R}'(\mathbf{x}) \to \mathsf{R}(\mathbf{x})$ into the rule set $\Sigma$.

**From Bounded Classes to the WA Class.** In this subsection, we show that any bounded ontology can be rewritten to a rule ontology that is weakly acyclic (Fagin et al. 2005).

Let us first review the notion of weak acyclicity. Fix $\Sigma$ as a set of rules and $\mathcal{R}$ its schema. A *position* of $\Sigma$ is a pair $(\mathsf{R}, i)$ where $\mathsf{R} \in \mathcal{R}$ is of an arity $n$ and $1 \leq i \leq n$. The *dependency graph* of $\Sigma$ is a directed graph with each position of $\Sigma$ as a node, and with each pair $((\mathsf{R}, i), (\mathsf{S}, j))$ as an edge if there is a rule $\varphi(\mathbf{x}, \mathbf{y}) \to \exists \mathbf{z} \psi(\mathbf{x}, \mathbf{z})$ from $\Sigma$ such that

- there is a variable $x \in \mathbf{x}$ such that $x$ occurs both in the position $(\mathsf{R}, i)$ in $\varphi$ and in the position $(\mathsf{S}, j)$ in $\psi$, or
- there are variables $x \in \mathbf{x}$ and $z \in \mathbf{z}$ such that $x$ occurs in the position $(\mathsf{R}, i)$ in $\varphi$ and $z$ occurs in the position $(\mathsf{S}, j)$ in $\psi$ (in this case, the edge is called a *special edge*).

A rule ontology $(\Sigma, \mathcal{D}, \mathcal{Q})$ is *weakly acyclic (WA)* if no cycle in the dependency graph of $\Sigma$ goes through a special edge.

It is well-known that the class of WA rule ontologies enjoys the finite chase property. In the last few years, a number of classes have been proposed to extend it. However, the next theorem shows that, in view of the expressiveness, the WA class is no weaker than any class with finite chase.

**Theorem 8.** *For every normal rule ontology $O = (\Sigma, \mathcal{D}, \mathcal{Q})$ with finite chase, there exists a weakly acyclic normal rule ontology $O^* = (\Sigma^*, \mathcal{D}, \mathcal{Q})$ such that $[\![O]\!] \approx [\![O^*]\!]$.*

We prove this theorem by developing a translation. The general idea is as follows. Given any normal rule ontology $O = (\Sigma, \mathcal{D}, \mathcal{Q})$ with finite chase, we need to construct a weakly acyclic rule ontology $O^* = (\Sigma^*, \mathcal{D}, \mathcal{Q})$ such that any computation on $O$ can be simulated by a computation on $O^*$. The main difficulty is how to simulate the cyclic existential quantifications by weakly acyclic existential quantifications. Fortunately, by Proposition 1, $O$ is always bounded, which means that the size of any (possibly functional) term generated by the chase procedure of $O$ on any database is bounded by some integer $\ell$. Therefore, every (possibly functional) term generated by the chase procedure of $O$ can be encoded by an $\ell$-tuple of function-free terms. For every relation symbol $\mathsf{R}$ occurring in $O$, we introduce an auxiliary relation symbol $\mathsf{R}^*$ with a larger arity so that every fact of the form $\mathsf{R}(\mathbf{t})$ can be encoded by a fact of the form $\mathsf{R}^*(\mathbf{t}^*)$.

With these settings, we can then construct some rules to simulate the chase procedure of $O$ by that of $O^*$ so that, for every $\mathcal{D}$-database $D$ and every fact $\mathsf{R}(\mathbf{t}) \in \mathsf{chase}(D, \Sigma)$, there exists a fact $\mathsf{R}^*(\mathbf{t}^*) \in \mathsf{chase}(D, \Sigma^*)$ that encodes $\mathsf{R}(\mathbf{t})$. These rules can be constructed by an approach similar to that in (Krötzsch and Rudolph 2011). So, the remaining task is to decode $\mathsf{R}(\mathbf{t})$ from $\mathsf{R}^*(\mathbf{t}^*)$. The difficulty of decoding is how to assure that, for each term (that might occur as arguments in different facts) in $\mathsf{chase}(D, \Sigma)$, there is exactly one null to be allocated to it. Fortunately again, it can be overcome by some encoding techniques. We will explain this later.

Now, let us define the translation formally. Let $\mathcal{D}, \mathcal{Q}$ and $\Sigma$ be defined as in the theorem. Let $\delta$ be a bound function such that $\Sigma$ is $\delta$-bounded. Let $m$ be the maximum arity of function symbols appearing in $\mathsf{sk}(\Sigma)$. Let $\ell = \sum_{i=0}^{\delta(\|\Sigma\|)} m^i$ and $s = (\ell - 1)/m$. Introduce a fresh $\ell \cdot n$-ary relation symbol[2] $\mathsf{R}^*$ for each $n$-ary relation symbol $\mathsf{R}$, introduce a fresh variable $v^i$ for each variable $v$ and each $i$ with $1 \leq i \leq \ell$, and introduce $\square$ (as a blank symbol to fill the gaps) and all non-nullary function symbols in $\mathsf{sk}(\Sigma)$ as fresh constants.

To represent a (functional) term $t$, we first regard $t$ as an $m$-complete tree with each symbol (function or constant) as a node and arguments of a function symbol $\mathsf{f}$ as the children of $\mathsf{f}$. If some node is empty, we then fill it with $\square$. The tuple encodes $t$ is then the symbol sequence obtained by a depth-first traversal. If the tuple is of length $< \ell$, we then fill $\square$s in the tail. Lastly, we let $[t]$ denote this tuple. For example, if $m = 2$, $\ell = 7$ and $t = \mathsf{f}(\mathsf{g}(a), b)$, then $[t] = \mathsf{fg}a\square b\square\square$.

To activate the simulation, some rules are needed to copy the data of $\mathsf{R}$ to $\mathsf{R}^*$. Formally, for each $n$-ary relation symbols $\mathsf{R} \in \mathcal{D}$, we need a rule $\varrho_\mathsf{R}$, defined as follows:

$$\mathsf{R}(x_1, \ldots, x_n) \to \mathsf{R}^*(x_1, \star, \cdots x_n, \star)$$

where $\star$ denotes the tuple consisting of $\ell - 1$ consecutive $\square$s.

Let $\gamma$ be a rule from $\Sigma$ of the form $\varphi(\mathbf{x}, \mathbf{y}) \to \exists \mathbf{z}\, \psi(\mathbf{x}, \mathbf{z})$ where $\mathbf{x} = x_1 \cdots x_k$ is a permutation of all the variables occurring in both $\varphi$ and $\psi$. We need a rule $\gamma^*$, which will be defined shortly, to simulate $\gamma$. For any term $t$ in $\gamma$, we let

$$\tau(t) := \begin{cases} a\square\cdots\square & \text{if } t = a \in \Delta; \\ v^1 \cdots v^s \square \cdots \square & \text{if } t = v \in \mathbf{x}; \\ \mathsf{f}_v x_1^1 \cdots x_1^s x_2^1 \cdots x_k^s \square \cdots \square & \text{if } t = v \in \mathbf{z}; \\ v^1 \cdots v^\ell & \text{if } t = v \in \mathbf{y}. \end{cases}$$

where, in each of the first three cases, the tail of $\tau(t)$ is filled with the symbol $\square$ such that the length of $\tau(t)$ is exactly $\ell$. Now we define $\gamma^*$ as the rule $\varphi^* \to \psi^*$, where $\varphi^*$ and $\psi^*$ denote the formulas obtained from $\varphi$ and $\psi$, respectively, by

---

[1] This is similar to that in data exchange (Fagin et al. 2005).

[2] In fact, we can use some relation symbol with a smaller arity, but this will make the presentation more complicated.

- substituting $\tau(t)$ for each term $t \in \Delta \cup \Delta_v$, followed by
- substituting $\mathsf{R}^*$ for each relation symbol $\mathsf{R}$.

In the chase procedure for the new ontology, by applying above rules on a $\mathcal{D}$-database $D$, we obtain a fact set $S^*$ that encodes $\mathsf{chase}(D, \Sigma)$. Thus, as mentioned previously, the remaining task is to construct rules for the decoding. The idea is as follows: (i) let $\mathsf{Dom}^*$ be the set of all $\ell$-tuples that encode terms with occurrences in $\mathsf{chase}(D, \Sigma)$; (ii) for each $\ell$-tuple $s^* \in \mathsf{Dom}^*$, generate a null $\mathsf{n}$ for it (by appying an existential quantifier once), and use $\mathsf{Map}(s^*, \mathsf{n})$ to record the correspondence between $s^*$ and $\mathsf{n}$; (iii) translate each fact $\mathsf{R}^*(t^*)$ to a fact $\mathsf{R}(t)$ by looking up the relation $\mathsf{Map}$.

To collect the $\ell$-tuples in stage (i), we need the following rules. Given an $n$-ary relation symbol $\mathsf{R} \in \mathcal{Q}$, let $\lambda_\mathsf{R}$ denote

$$\mathsf{R}^*(\boldsymbol{v}_1, \ldots, \boldsymbol{v}_n) \to \mathsf{Dom}^*(\boldsymbol{v}_1) \wedge \cdots \wedge \mathsf{Dom}^*(\boldsymbol{v}_n)$$

where each $\boldsymbol{v}_i$ is a tuple of distinct variables $v_i^1 \cdots v_i^s$, and $\mathsf{Dom}^*$ a fresh relation symbol of arity $\ell$.

Next, we define some rules to generate nulls, which implement stage (ii). For each function symbol $\mathsf{f}_x$ in $\mathsf{sk}(\Sigma)$ where $x$ is an existential variable in $\Sigma$, let $\zeta_x$ denote

$$\mathsf{Dom}^*(\mathsf{f}_x, \boldsymbol{v}) \to \exists x\, \mathsf{Map}(\mathsf{f}_x, \boldsymbol{v}, x)$$

where $\boldsymbol{v}$ is a tuple of distinct variables $v_1 \cdots v_{\ell-1}$, and $\mathsf{Map}$ a fresh $(\ell+1)$-ary relation symbol. In addition, let $\zeta_c$ denote

$$\mathsf{Dom}^*(x, \square, \ldots, \square) \to \mathsf{Map}(x, \square, \ldots, \square, x).$$

Informally, this rule asserts that, for any $\ell$-tuple that encodes a single-symbol term, we do not need to generate any null.

Now, we can define rules to carry out the decoding. For each $n$-ary relation symbol $\mathsf{R} \in \mathcal{Q}$, let $\vartheta_\mathsf{R}$ denote

$$\begin{bmatrix} \mathsf{R}^*(\boldsymbol{v}_1, \ldots, \boldsymbol{v}_n) \wedge \mathsf{Map}(\boldsymbol{v}_1, x_1) \\ \wedge \cdots \wedge \mathsf{Map}(\boldsymbol{v}_n, x_n) \end{bmatrix} \to \mathsf{R}(x_1, \ldots, x_n).$$

Finally, we let $\Sigma^*$ denote the rule set consisting of (1) $\varrho_\mathsf{R}$ for every relation symbol $\mathsf{R} \in \mathcal{D}$, (2) $\gamma^*$ for every rule $\gamma \in \Sigma$, (3) $\lambda_\mathsf{R}$ for every relation symbol $\mathsf{R} \in \mathcal{Q}$, (4) $\zeta_x$ for every existential variable $x$ in $\Sigma$ such that $\mathsf{f}_x$ is of a positive arity, (5) $\zeta_c$, and (6) $\vartheta_\mathsf{R}$ for every relation symbol $\mathsf{R} \in \mathcal{Q}$.

**Example 2.** By adding a copy rule into the rule ontology $O$ defined in Example 1, we then obtain a normal rule ontology $O_0 = (\Sigma_0, \mathcal{D}_0, \mathcal{Q}_0)$, where $\Sigma_0$ is the following rule set:

$$\mathsf{D}(x, y) \to \mathsf{R}(x, y)$$
$$\mathsf{R}(x, x) \to \exists yz\, \mathsf{S}(x, y) \wedge \mathsf{S}(y, z)$$
$$\mathsf{R}(x, y) \wedge \mathsf{S}(x, z) \to \exists v\, \mathsf{R}(z, v)$$

and $\mathcal{D}_0 = \{\mathsf{D}\}$, $\mathcal{Q}_0 = \{\mathsf{R}\}$. Next, we will illustrate the translation by the rule ontology $O_0$.

All the function symbols in $\mathsf{sk}(\Sigma_0)$ are clearly unary, and as analyzed in Example 1, $O_0$ is $\delta$-bounded for some bound function $\delta$ with $\delta(\|\Sigma_0\|) = 2$. So, we have $\ell = 1^0 + 1^1 + 1^2 = 3$, i.e., terms generated by the chase procedure of $O_0$ will be encoded by triples of function-free terms. Now, we use the following rule to initialize the auxiliary relation symbol $\mathsf{D}^*$:

$$\mathsf{D}(x, y) \to \mathsf{D}^*(x \square \square\, y \square \square)$$

To simulate the chase procedure of $O_0$, we need the following rules, which correspond to the rules in $\Sigma_0$:

$$\mathsf{D}^*(x_1 x_2 x_3\, y_1 y_2 y_3) \to \mathsf{R}^*(x_1 x_2 x_3\, y_1 y_2 y_3)$$
$$\mathsf{R}^*(x_1 x_2 \square\, x_1 x_2 \square) \to \mathsf{S}^*(x_1 x_2 \square\, \mathsf{f}_y x_1 x_2) \wedge \mathsf{S}^*(\mathsf{f}_y x_1 x_2\, \mathsf{f}_z x_1 x_2)$$
$$\mathsf{R}^*(x_1 x_2 x_3\, y_1 y_2 y_3) \wedge \mathsf{S}^*(y_1 y_2 y_3\, z_1 z_2 \square) \to \mathsf{R}^*(z_1 z_2 \square\, \mathsf{f}_v z_1 z_2)$$

The following rules are used to implement the decoding:

$$\mathsf{R}^*(x_1 x_2 x_3\, y_1 y_2 y_3) \to \mathsf{Dom}^*(x_1 x_2 x_3) \wedge \mathsf{Dom}^*(y_1 y_2 y_3)$$
$$\mathsf{Dom}^*(\mathsf{f}_y x_1 x_2) \to \exists y\, \mathsf{Map}(\mathsf{f}_y x_1 x_2\, y)$$
$$\mathsf{Dom}^*(\mathsf{f}_z x_1 x_2) \to \exists z\, \mathsf{Map}(\mathsf{f}_z x_1 x_2\, z)$$
$$\mathsf{Dom}^*(\mathsf{f}_v x_1 x_2) \to \exists v\, \mathsf{Map}(\mathsf{f}_v x_1 x_2\, v)$$
$$\mathsf{Dom}^*(x \square \square) \to \mathsf{Map}(x \square \square\, x)$$
$$\begin{bmatrix} \mathsf{R}^*(x_1 x_2 x_3\, y_1 y_2 y_3) \wedge \\ \mathsf{Map}(x_1 x_2 x_3\, x) \wedge \mathsf{Map}(y_1 y_2 y_3\, y) \end{bmatrix} \to \mathsf{R}(x\, y)$$

Finally, let $\Sigma_0^*$ consist of the set of all rules defined above. It is not difficult to check that $[\![O_0]\!] \approx [\![(\Sigma_0^*, \mathcal{D}_0, \mathcal{Q}_0)]\!]$. □

**Capturing PTIME by the WA Class.** We have proved that all the rule languages with finite chase are of the same expressiveness as the WA class in the last subsection. However, this characterization is syntactic. In this subsection, we will give a complexity-theoretic characterization. Before presenting the result, we need some definitions.

Like in traditional Datalog (Dantsin et al. 2001), we will study the expressiveness on ordered databases. Every *ordered database* is a database whose domain is an integer set $\{0, \ldots, n\}$ for some integer $n \geq 0$; whose schema contains three special relation symbols $\mathsf{Succ}$, $\mathsf{Min}$ and $\mathsf{Max}$ (we call such a schema *ordered*); in which $\mathsf{Succ}$ is interpreted as the immediate successor relation on natural numbers, and $\mathsf{Min}$ and $\mathsf{Max}$ are interpreted as $\{0\}$ and $\{n\}$, respectively. By *ordered UWMs* we mean the restrictions of UWMs to ordered databases. We generalize definitions of $[\![\cdot]\!]$ and $\approx$ to ordered UWMs by replacing "database" with "ordered database". Note that the ordered version of Proposition 7 still holds.

We fix a natural way for representing (general) databases in binary strings. Given a general database $D$, let $\langle D \rangle$ denote its binary represention. Let $\mathcal{D}$ and $\mathcal{Q}$ be any two disjoint relational schemas where $\mathcal{D}$ is ordered. Let $\Phi$ be an ordered UWM over $(\mathcal{D}, \mathcal{Q})$. We say that $\Phi$ is *computed* by a Turing machine $M$ if $M$ halts on any input $\langle D \rangle$ where $D$ is an ordered $\mathcal{D}$-database, and there is a general $\mathcal{Q}$-database $Q$ such that $Q$ is homomorphically equivalent to $\Phi(D)$ and the output w.r.t. $\langle D \rangle$ is $\langle Q \rangle$, the binary representation of $Q$.

On syntax, we also need a slightly richer language defined as follows. Let $\mathcal{D}$ be a relational schema (as a database schema). A *semipositive rule* w.r.t. $\mathcal{D}$ is a generalized rule defined by allowing negated atoms with relation symbols from $\mathcal{D}$ to appear in the body. Semipositive rule ontologies are then generalized from rule ontologies by allowing semipositive rules w.r.t. its database schema. A semipositive rule ontology is called *weakly acyclic* if the rule ontology obtained by omitting negative atoms is weakly acyclic.

**Theorem 9.** *For every ordered UWM $\Phi$ that is computable in deterministic polynomial time, there is a weakly acyclic and semipositive rule ontology $O$ such that $[\![O]\!] \approx \Phi$.*

*Remark* 4. By a slight generalization of the critical database technique proposed in (Marnette 2009), one can show that every semipositive rule ontology with finite Skolem chase is computable in deterministic polynomial time. Therefore, the above theorem implies that every semipositive rule language with finite Skolem chase that extends the semipositive WA class captures the class of PTIME-computable UWMs.

**Succinctness.** Our previous results show that all the rule languages with finite chase that extend the weakly acyclic class are of the same expressiveness. Now we further consider the following question: Is it possible to compare the efficiency of rule languages with finite chase for representing ontological knowledge? In general, it is not an easy task to compare the succinctness for fragments of first-order logic. However, the following theorem provides us with such a result for rule languages, which states that the bounded rule languages with higher combined complexity are normally more succinct than those with lower combined complexity.

**Theorem 10.** *For all $k > 0$, there exists a $k$-exponentially bounded rule ontology $O = (\Sigma, \mathcal{D}, \mathcal{Q})$ such that, for any $(k-1)$-exponentially bounded rule ontology $O_0 = (\Sigma_0, \mathcal{D}, \mathcal{Q})$ where $\Sigma_0$ is of polynomial size w.r.t. $\Sigma$, we have $[\![O_0]\!] \not\approx [\![O]\!]$.*

*Proof.* (Sketch) Let $n, \ell$ and $\Sigma_{\mathsf{num}}$ be defined as in the proof of Proposition 4. Let $\mathcal{D} = \emptyset$ and $\mathcal{Q} = \{\mathsf{Min}_\ell, \mathsf{Max}_\ell, \mathsf{Succ}_\ell\}$. By using the notion of *core* (see, e.g., (Deutsch, Nash, and Remmel 2008)), we show a lower bound for the number of nulls in universal models. By estimating the number of nulls to be used in the chase, we then prove that $(\Sigma_{\mathsf{num}}, \mathcal{D}, \mathcal{Q})$ is not equivalent to any $(k-1)$-exponentially bounded ontology $(\Sigma, \mathcal{D}, \mathcal{Q})$ if $\Sigma$ is of a polynomial size w.r.t. $\Sigma_{\mathsf{num}}$. □

*Remark* 5. Theorem 10 tells us that, although extending the WA class to larger classes with finite chase does not increase the expressiveness, the succinctness could be a bonus.

*Remark* 6. It would be interesting to compare the succinctness of finite-chase rule languages with the same combined complexity under query answering. For instance, is the MFA class more succinct than the WA class? But this is beyond the scope of this work. We will pursue it in the future.

## Concluding Remarks

We have studied the existential rule languages with finite chase in this paper. Instead of considering specific rule languages like most current works on this topic, here we have defined a family of rule languages based on a new concept called $\delta$-*boundedness*, from which the overall complexity and expressiveness characterizations on these languages have been provided. Our study on this topic may be further undertaken in various directions. One interesting yet challenging future work is to investigate disjunctive existential rule languages. It is important to discover whether our approach can be extended to identify decidable disjunctive existential rule languages and to characterize relevant complexity and expressiveness properties. Results on this aspect may significantly enhance our current understanding on ontological reasoning with disjunctive existential rules (Alviano et al. 2012; Bourhis, Morak, and Pieris 2013).

# Appendix: Detailed Proofs

## Proof of Proposition 1

**Proposition 1.** There is a bound function $\delta$ such that, for every rule ontology $O$, $O$ has finite chase iff it is $\delta$-bounded.

To show this, we use a technique developed by (Marnette 2009). Let $O = (\Sigma, \mathcal{D}, \mathcal{Q})$ be a rule ontology. Let $C$ denote the set of constants appearing in $\Sigma$ and let $*$ be a special constant without occurrence in $\Sigma$. A database over $\mathcal{D}$ is called *critical* for $O$ if each relation in it is a full relation on the domain $C \cup \{*\}$. Clearly, the critical database for $O$ is unique. For convenience, let $D_O^*$ denote the critical database of $O$.

*Proof of Proposition 1.* Given an arbitrary bound function $\delta$, it is clear that every $\delta$-bounded rule ontology has finite chase. So, it suffices to show that there is a bound function $\delta$ such that every rule ontology with finite chase is $\delta$-bounded. To do this, we first need to prove a claim as follow.

**Claim 1.** *Let* $O = (\Sigma, \mathcal{D}, \mathcal{Q})$ *be a rule ontology and $D$ a $\mathcal{D}$-database. Then* $\mathsf{ht}(\mathsf{chase}(D, \Sigma)) \leq \mathsf{ht}(\mathsf{chase}(D_O^*, \Sigma))$.

*Proof.* Let $f$ denote the function that maps every constant in $\Delta$ to itself if it appears in $\Sigma$, and $*$ otherwise. Furthermore, we generalize $f$ to terms, atoms and general instances in the standard way. By a routine induction, one can easily show that $f(\mathsf{chase}^n(D, \Sigma)) \subseteq \mathsf{chase}^n(D_O^*, \Sigma)$ for all $n \geq 0$, which implies $f(\mathsf{chase}(D, \Sigma)) \subseteq \mathsf{chase}(D_O^*, \Sigma)$ immediately. Since $\mathsf{ht}(f(t)) = \mathsf{ht}(t)$ for every term $t$, we can then conclude that $\mathsf{ht}(\mathsf{chase}(D, \Sigma)) \leq \mathsf{ht}(\mathsf{chase}(D_O^*, \Sigma))$. □

Now we are in the position to complete the proof. Let $p$ be a function that maps each rule ontology $O = (\Sigma, \mathcal{D}, \mathcal{Q})$ with finite chase to $\mathsf{ht}(\mathsf{chase}(D_O^*, \Sigma))$. It is clear that $p(O)$ is a positive integer for any rule ontology $O$ with finite chase. Now, for all $n > 0$ let $\delta(n)$ be the maximal $p(O)$ for all rule ontologies $O = (\Sigma, \mathcal{D}, \mathcal{Q})$ with finite chase and $\|\Sigma\| = n$. Since the number of rule sets of size $n$ is finite, $\delta$ should be a bound function. By Claim 1, every rule ontology with finite chase is clearly $\delta$-bounded, which is as desired. □

## Proof of Proposition 2

**Proposition 2.** There is no computable bound function $\delta$ such that every bounded rule ontology is $\delta$-bounded.

*Proof.* Towards a contradiction, we assume that such a computable function $\delta$ exists. Let $\Sigma$ be any rule ontology. Then, by Remark 1, to check whether $\Sigma$ has finite chase, it is equivalent to check whether $\Sigma$ is $\delta$-bounded. Since $\delta$ is computable, by Proposition 6 (whose proof is given shortly in this appendix), there is an algorithm to check whether $\Sigma$ is $\delta$-bounded. This means that the problem of checking whether a rule ontology has finite Skolem chase is decidable, which contradicts with Theorem 4 in (Marnette 2009), which states that it is RE-complete to decide whether a tg schema-mapping (i.e., rule ontology) has finite Skolem chase. This completes the proof. □

## Proof of Proposition 3

**Proposition 3.** Let $\delta$ be a bound function. Then for any $\delta$-bounded rule ontology $(\Sigma, \mathcal{D}, \mathcal{Q})$, any $\mathcal{D}$-database $D$ and any BCQ $q$ over $\mathcal{Q}$, it is in

$$\mathrm{DTIME}((|\mathsf{dom}(D)| + \|\Sigma\|)^{\|\Sigma\|^{\mathcal{O}(\delta(\|\Sigma\|))} \cdot \|q\|^{\mathcal{O}(1)}})$$

to check whether $D \cup \Sigma \models q$.

*Proof.* Let $(\Sigma, \mathcal{D}, \mathcal{Q})$ be a $\delta$-bounded rule ontology, $D$ a $\mathcal{D}$-database, and $q$ a BCQ over $\mathcal{Q}$. Let $k$, $n$, $\ell$, $m$, and $c$ denote the number of relation symbols appearing in $\Sigma$, the maximal arity of relation symbols appearing in $\Sigma$, the number of function symbols in $\mathsf{sk}(\Sigma)$, the maximal arity of function symbols appearing in $\mathsf{sk}(\Sigma)$, and the number of constants appearing in $D$, respectively. Let $\mathsf{R}(\boldsymbol{t})$ be any fact in $\mathsf{chase}(D, \Sigma)$. By the definition of $\delta$-boundedness, it is clear that every component $t \in \boldsymbol{t}$ contains at most $\sum_{i=0}^{\delta(\|\Sigma\|)} m^i = m^{\mathcal{O}(\delta(\|\Sigma\|))}$ symbols, and each symbol is either a constant or a function symbol. So, $\mathsf{chase}(D, \Sigma)$ consists of $(c+\ell)^{m^{\mathcal{O}(\delta(\|\Sigma\|))} \cdot n} \cdot k$ facts. Since $k, n, \ell, m \leq \|\Sigma\|$ and $c = |\mathsf{dom}(D)|$, the chase on $D$ and $\Sigma$ must terminate in $(|\mathsf{dom}(D)| + \|\Sigma\|)^{\|\Sigma\|^{\mathcal{O}(\delta(\|\Sigma\|))} \cdot \|\Sigma\|^{\mathcal{O}(1)}}$ steps. It is also clear that each step of the chase can be computed in $\mathrm{DTIME}((|\mathsf{dom}(D)| + \|\Sigma\|)^{\|\Sigma\|^{\mathcal{O}(\delta(\|\Sigma\|))}})$. Thus, $\mathsf{chase}(D, \Sigma)$ can be computed in $\mathrm{DTIME}((|\mathsf{dom}(D)| + \|\Sigma\|)^{\|\Sigma\|^{\mathcal{O}(\delta(\|\Sigma\|))}})$, too.

To complete the query answering, it is now sufficient to evaluate $q$ on $\mathsf{chase}(D, \Sigma)$ directly. Without loss of generality, we assume that $q$ is in prenex normal form. Let $s$ be the number of existential variables occurring in $q$. To evaluate the chase, it is equivalent to check whether there is a substitution $h$, mapping every existential variable to a ground term of height less than $\delta(\|\Sigma\|)$, such that $h(q) \subseteq \mathsf{chase}(D, \Sigma)$. By the previous analysis, there are at most $(|\mathsf{dom}(D)| + \|\Sigma\|)^{\|\Sigma\|^{\mathcal{O}(\delta(\|\Sigma\|))} \cdot s}$ substitutions that need to be check. In addition, it is in $\mathrm{DTIME}((|\mathsf{dom}(D)| + \|\Sigma\|)^{\|\Sigma\|^{\mathcal{O}(\delta(\|\Sigma\|))} \cdot \|q\|^{\mathcal{O}(1)}})$ to check whether $h(q) \subseteq \mathsf{chase}(D, \Sigma)$. Since $s \leq \|q\|$, the evaluation can then be finished in $\mathrm{DTIME}((|\mathsf{dom}(D)| + \|\Sigma\|)^{\|\Sigma\|^{\mathcal{O}(\delta(\|\Sigma\|))} \cdot \|q\|^{\mathcal{O}(1)}})$. Combining it with the result in previous paragraph, we then have the desired proposition. □

## Proof of Theorem 5

**Theorem 5.** For all integers $k \geq 0$, the Boolean query answering problem of the $k$-exponentially bounded language is $(k+2)$-EXPTIME-complete for the combined complexity, and PTIME-complete for the data complexity.

*Proof.* The combined complexity is by Propositions 3 and 4. The membership of the data complexity is by Proposition 3 (also in fact implied by Theorem 3 of (Marnette 2009)), and the hardness follows from the PTIME-completeness of data complexity for Datalog, see, e.g., (Dantsin et al. 2001). □

## Proof of Proposition 6

**Proposition 6.** Let $\delta$ be a bound function that is computable in $\mathrm{DTIME}(T(n))$ for some function $T(n)$. Then for every rule ontology $O = (\Sigma, \mathcal{D}, \mathcal{Q})$, it is in

$$\mathrm{DTIME}(\|\Sigma\|^{\|\Sigma\|^{\mathcal{O}(\delta(\|\Sigma\|))}} + T(\log \|\Sigma\|)^{\mathcal{O}(1)})$$

to check whether $O$ is $\delta$-bounded.

*Proof.* Let $O = (\Sigma, \mathcal{D}, \mathcal{Q})$ be a rule ontology. According to Claim 1, we can infer that $O$ is $\delta$-bounded if and only if $\mathsf{ht}(\mathsf{chase}(D_O^*, \Sigma)) \leq \delta(\|\Sigma\|)$, where $D_O^*$ is the critical database defined as previous. So, to check if $O$ is $\delta$-bounded, it is equivalent to check if $\mathsf{ht}(\mathsf{chase}(D_O^*, \Sigma)) \leq \delta(\|\Sigma\|)$. By the analysis in the proof of Proposition 3, if the inequality holds, $\mathsf{chase}(D_\Sigma^*, \Sigma)$ should be of the size $\|\Sigma\|^{\|\Sigma\|^{\mathcal{O}(\delta(\|\Sigma\|))}}$. In particular, an upper bound for this size can be computed in $\mathsf{DTIME}((\|\Sigma\| + T(\log\|\Sigma\|))^{\mathcal{O}(1)})$. Let $n$ denote the resulting upper bound. Now, we can design an algorithm to simulate the first $n+1$ stages of the chase on $D_O^*$ and $\Sigma$. If $\mathsf{ht}(\mathsf{chase}^{n+1}(D_O^*, \Sigma)) \leq \delta(\|\Sigma\|)$, then $\Sigma$ should be $\delta$-bounded, otherwise not. It is clear that the full computation can be implemented in $\mathsf{DTIME}(\|\Sigma\|^{\|\Sigma\|^{\mathcal{O}(\delta(\|\Sigma\|))}} + T(\log\|\Sigma\|)^{\mathcal{O}(1)})$. $\square$

## Proof of Proposition 7

**Proposition 7.** *Let $O_1 = (\Sigma_1, \mathcal{D}, \mathcal{Q})$ and $O_2 = (\Sigma_2, \mathcal{D}, \mathcal{Q})$ be two rule ontologies with finite chase. Then $[\![O_1]\!] \approx [\![O_2]\!]$ iff, for all $\mathcal{D}$-databases $D$ and all BCQs $q$ over $\mathcal{Q}$, we have*

$$D \cup \Sigma_1 \models q \quad \text{iff} \quad D \cup \Sigma_2 \models q.$$

*Proof.* The direction of "only-if" is trivial. We only show the converse. Assume that $D \cup \Sigma_1 \models q$ if and only if $D \cup \Sigma_2 \models q$ for all $\mathcal{D}$-databases $D$ and all BCQs $q$ over $\mathcal{Q}$. Let $D$ be any $\mathcal{D}$-database. For $i=1$ or 2, let $q_i$ denote the BCQ obtained from $\mathsf{chase}(D, \Sigma_i)$ by replacing each null (i.e., a functional term) by a fresh existential variable. Note that both $O_1$ and $O_2$ have finite chase, so such BCQs exist. Then it is clear that $D \cup \Sigma_i \models q_i$. Thus, by the assumption we have that $\mathsf{chase}(D, \Sigma_2) \models q_1$ and $\mathsf{chase}(D, \Sigma_1) \models q_2$. From these, we can infer that $\mathsf{chase}(D, \Sigma_1)$ and $\mathsf{chase}(D, \Sigma_1)$ are homomorphically equivalent. This completes the proof. $\square$

## Proof of Theorem 8

**Theorem 8.** *For every normal rule ontology $O = (\Sigma, \mathcal{D}, \mathcal{Q})$ with finite chase, there exists a weakly acyclic normal rule ontology $O^* = (\Sigma^*, \mathcal{D}, \mathcal{Q})$ such that $[\![O]\!] \approx [\![O^*]\!]$.*

*Proof.* Let $\mathcal{R}$ be the schema of $\Sigma$, and let $\mathcal{R}^*$ consist of $\mathsf{R}^*$ for all relation symbols $\mathsf{R} \in \mathcal{R}$. Given a ground term $t$, let $[t]$ be defined as previous (after the statement of Theorem 8). Now, we need to generalized the mapping $[\cdot]$ to atoms and general instances. For any atom $\alpha$ of the form $\mathsf{R}(t_1, \ldots, t_n)$, we define $[\alpha] = \mathsf{R}^*([t_1], \ldots, [t_n])$. Given any general instance $I$, let $[I]$ denote the set of $[\alpha]$ for all $\alpha \in I$. To prove the desired theorem, we need two claims.

**Claim 2.** $[\mathsf{chase}^n(D, \Sigma)] \subseteq \mathsf{chase}^{n+1}(D, \Sigma^*)|_{\mathcal{R}^*}$ *for $n \geq 0$.*

*Proof of Claim 2.* We show this by an induction on $n$. It is clear for $n=0$ since $\mathsf{chase}^0(D, \Sigma) = D$ and any fact from $[D]$ can be obtained from $D$ by applying rules $\varrho_R$ in one chase stage. Assume $n > 0$, and suppose as inductive hypothesis that $[\mathsf{chase}^{n-1}(D, \Sigma)] \subseteq \mathsf{chase}^n(D, \Sigma^*)|_{\mathcal{R}^*}$. Let $\alpha \in [\mathsf{chase}^n(D, \Sigma)]$ be an atom of form $[\mathsf{R}(t_1, \ldots, t_k)]$ for some $\mathsf{R} \in \mathcal{R}$. Then it is clear that $\mathsf{R}(t_1, \ldots, t_k) \in \mathsf{chase}^n(D, \Sigma)$. By the definition of chase, there exist a rule $\gamma \in \Sigma$ and a substitution $h$ such that $h(\mathsf{body}(\gamma)) \subseteq \mathsf{chase}^{n-1}(D, \Sigma)$ and $\mathsf{R}(t_1, \ldots, t_k) \in h(\mathsf{head}(\gamma))$. By the inductive hypothesis, we can conclude that $[h(\mathsf{body}(\gamma))] \subseteq \mathsf{chase}^n(D, \Sigma^*)$. Let $[h]$ be the substitution that maps each variable $x$ to $[h(x)]$. We then have that $[h](\mathsf{body}(\gamma^*)) \subseteq \mathsf{chase}^n(D, \Sigma^*)$. Consequently, it holds that $[h](\mathsf{head}(\gamma^*)) \subseteq \mathsf{chase}^{n+1}(D, \Sigma^*)$, or equivalently $[h(\mathsf{head}(\gamma))] \subseteq \mathsf{chase}^{n+1}(D, \Sigma^*)$. Thus, we have $\alpha \in \mathsf{chase}^{n+1}(D, \Sigma^*)$. This yields the desired claim. $\square$

**Claim 3.** $[\mathsf{chase}^n(D, \Sigma)] \supseteq \mathsf{chase}^n(D, \Sigma^*)|_{\mathcal{R}^*}$ *for $n \geq 0$.*

*Proof of Claim 3.* Again, we show this by an induction on $n$. It is clearly true for $n=0$ since $\mathsf{chase}^0(D, \Sigma^*)|_{\mathcal{R}^*} = \emptyset$. Now we assume $n > 0$ and suppose as inductive hypothesis that $[\mathsf{chase}^{n-1}(D, \Sigma)] \supseteq \mathsf{chase}^{n-1}(D, \Sigma^*)|_{\mathcal{R}^*}$. Let $\alpha^*$ be any atom from $\mathsf{chase}^n(D, \Sigma^*)$. By the definition of chase, there must exist a rule $\gamma_0 \in \Sigma^*$ and an assignment $h^*$ such that $h^*(\mathsf{body}(\gamma_0)) \subseteq \mathsf{chase}^{n-1}(D, \Sigma^*)$ and $\alpha^* \in h^*(\mathsf{head}(\gamma_0))$. So it suffices to show $\alpha^* \in [\mathsf{chase}^n(D, \Sigma)]$ for each of the following cases: (1) $\gamma_0 = \varrho_\mathsf{R}$ for some $\mathsf{R} \in \mathcal{D}$; (2) $\gamma_0 = \gamma^*$ for some $\gamma \in \Sigma$. Let us first assume case (1), and suppose that $\alpha^*$ is of form $\mathsf{R}^*(c_1, \star, \cdots, c_k, \star)$ for some $k$-ary relation symbol $\mathsf{R} \in \mathcal{D}$, where $\star$ denotes the tuple consisting of $\ell - 1$ consecutive $\square$s. Then, it is not difficult to see that $\mathsf{R}(c_1, \ldots, c_k) \in D$, which implies $\alpha^* \in [\mathsf{chase}^n(D, \Sigma)]$ immediately. Next, let us consider case (2). Let $\gamma$ be the rule from $\Sigma$ such that $\gamma_0 = \gamma^*$. Let $h$ be an assignment that maps each variable $x$ to $[h^*(x)]^{-1}$, where $[\cdot]^{-1}$ denotes the inverse function of $[\cdot]$. (Clearly, such a function exists.) Then, by the definition of $\gamma^*$, it is clear that $[\alpha^*]^{-1} \in h(\mathsf{head}(\gamma))$. To show $\alpha^* \in [\mathsf{chase}^n(D, \Sigma)]$, it is enough to show $h(\mathsf{body}(\gamma)) \subseteq \mathsf{chase}^{n-1}(D, \Sigma)$, which can be obtained from the definition of $\gamma^*$ and the inductive hypothesis. $\square$

By these two claims, we then have that $[\mathsf{chase}(D, \Sigma)] = \mathsf{chase}(D, \Sigma^*)|_{\mathcal{R}^*}$. On the other hand, let $g$ be a function that maps each variable-free $\ell$-tuple $\boldsymbol{t}$ to the first component of $\boldsymbol{t}$ if the second component of $\boldsymbol{t}$ is $\square$, and $\mathsf{f}_x(\boldsymbol{t})$ (where $x$ is a variable) otherwise. By rules $\zeta_x$ and $\zeta_c$, we see that $\mathsf{Map}$ is the graph of $g$. Let $h$ be the function that maps each term $t$ to $g([t])$. It is not difficult to check that $h$ is a homomorphism from $\mathsf{chase}(D, \Sigma)|_\mathcal{Q}$ to $\mathsf{chase}(D, \Sigma^*)|_\mathcal{Q}$ and $h^{-1}$ is a homomorphism from $\mathsf{chase}(D, \Sigma^*)|_\mathcal{Q}$ to $\mathsf{chase}(D, \Sigma)|_\mathcal{Q}$. $\square$

## Proof of Theorem 9

**Theorem 9.** *For every ordered UWM $\Phi$ that is computable in deterministic polynomial time, there is a weakly acyclic and semipositive rule ontology $O$ such that $[\![O]\!] \approx \Phi$.*

*Proof.* Let $\Phi$ be a UWM for ordered database that is computed by a deterministic Turing machine $M$ in polynomial time, where $\mathcal{D}$ and $\mathcal{Q}$ are disjoint relational schemas. Then there is an integer $k \geq 0$ such that $M$ will halt on every ordered $\mathcal{D}$-database $D$ in $|\mathsf{dom}(D)|^k$ steps. To show the theorem, it is sufficient to construct a weakly acyclic and semipositive rule ontology $O = (\Sigma, \mathcal{D}, \mathcal{Q})$ such that $[\![O]\!] \approx \Phi$. W.l.o.g., we assume the $M$ has only three tape symbols: "0", "1" and "$\diamond$" (the blank symbol), and a one-way tape in which cells are indexed by natural numbers. Both the input and output are stored on the tape started from the 0-th cell.

Let $D$ be any ordered $\mathcal{D}$-database. Again, the first task is to define a linear order of length $\geq |\mathsf{dom}(D)|^k$. Clearly, we can use the following rule to assert that the unary relation Const consists of all the constants from the domain of $D$:

$$\mathsf{Succ}(x, y) \to \mathsf{Const}(x) \land \mathsf{Const}(y).$$

With it, we can then generate $|\mathsf{dom}(D)|^k$ new elements by

$$\mathsf{Const}(x_1) \land \cdots \mathsf{Const}(x_k) \to \exists y\, \mathsf{G}(x_1, \ldots, x_k, y)$$

where new elements are stored in the last argument of G. To define a linear order (it can be built from the original order Succ) on these new elements, we need the following rule

$$\begin{bmatrix} \mathsf{G}(\boldsymbol{x}_i, y, \boldsymbol{z}_i, u) \land \mathsf{G}(\boldsymbol{x}_i, y_0, \boldsymbol{v}_i, w) \\ \land \mathsf{Succ}(y, y_0) \land \mathsf{Max}(z) \land \mathsf{Min}(v) \end{bmatrix} \to \mathsf{Succ}^*(u, w)$$

for each $i$ with $1 \leq i \leq k$, where $\boldsymbol{x}_i$ is an $(i-1)$-tuple of distinct variables, and $\boldsymbol{z}_i$ and $\boldsymbol{v}_i$ denote the $(k-i-1)$-tuples $z \cdots z$ and $v \cdots v$, respectively. Clearly, $\mathsf{Succ}^*$ defines a linear order on new elements (the lexicographical order generated from Succ). For technical reasons, we will combine it with the old order. This can be done by the following rules:

$$\mathsf{Min}(x) \to \mathsf{Min}^*(x)$$
$$\mathsf{Succ}(x, y) \to \mathsf{Succ}^*(x, y)$$
$$\mathsf{Max}(x) \land \mathsf{Min}(y) \land \mathsf{G}(y, \ldots, y, z) \to \mathsf{Succ}^*(x, z)$$
$$\mathsf{Max}(x) \land \mathsf{G}(x, \ldots, x, y) \to \mathsf{Max}^*(y)$$

Now, by applying the above rules, we then have a linear order of length $|\mathsf{dom}(D)| + |\mathsf{dom}(D)|^k$. To complete the construction, we still need to define some arithmetical relations. This can be done in a routine way, e.g., we can define the relation Add of addition by the following rules:

$$\mathsf{Succ}^*(x, y) \to \mathsf{Num}(x) \land \mathsf{Num}(y)$$
$$\mathsf{Num}(x) \land \mathsf{Min}^*(y) \to \mathsf{Add}(x, y, x)$$
$$\mathsf{Add}(x, y, z) \land \mathsf{Succ}^*(y, u) \land \mathsf{Succ}^*(z, v) \to \mathsf{Add}(x, u, v)$$

In addition, we let $\mathsf{LE}^*$ define the relation "less than or equal to". It can be defined by the following rules:

$$\mathsf{Num}(x) \to \mathsf{LE}^*(x, x)$$
$$\mathsf{Succ}^*(x, y) \to \mathsf{LE}^*(x, y)$$
$$\mathsf{LE}^*(x, y) \land \mathsf{LE}^*(y, z) \to \mathsf{LE}^*(x, z)$$

With the linear order and arithmetical relations, we are then in the position to represent the Turing machine $M$.

Without loss of generality, we assume $\mathcal{D} = \{\mathsf{D}_0, \ldots \mathsf{D}_d\}$ and $\mathcal{Q} = \{\mathsf{Q}_0, \ldots \mathsf{Q}_q\}$ for some $d, q > 0$. We let $\mathsf{DPos}_i(\boldsymbol{x}, v)$ be a binary relation asserting that the truth of fact $\mathsf{D}_i(\boldsymbol{x})$ has been stored in the $v$-th tape cell initially; $\mathsf{QPos}_i(\boldsymbol{x}, v)$ be a binary relation asserting that the truth of fact $\mathsf{Q}_i(\boldsymbol{x})$ should be stored in the $v$-th tape cell finally; $\mathsf{DSize}(x)$ and $\mathsf{QSize}(y)$ asserting that the sizes of input string and output string are $x$ and $y$, respectively. It is not difficult to see that, for a natural encoding approach of the (general) database, these relations can be defined by using some arithmetical relations.

To set the initial configuration, we need following rules

$$\mathsf{Min}^*(x) \to \mathsf{Head}(x, x) \land \mathsf{State}(x, s_0)$$
$$\mathsf{D}_i(\boldsymbol{v}) \land \mathsf{DPos}_i(\boldsymbol{v}, x) \land \mathsf{Min}^*(y) \to \mathsf{Tape}(y, x, 1)$$
$$\neg \mathsf{D}_i(\boldsymbol{v}) \land \mathsf{DPos}_i(\boldsymbol{v}, x) \land \mathsf{Min}^*(y) \to \mathsf{Tape}(y, x, 0)$$
$$\mathsf{DSize}(z) \land \mathsf{LE}^*(z, x) \land \mathsf{Min}^*(y) \to \mathsf{Tape}(y, x, \diamond)$$

for each relation symbol $\mathsf{D}_i \in \mathcal{D}$, where $\mathsf{Tape}(x, y, z)$ states that, in time $x$, the tape symbol $z$ is written on the $y$-th tape cell, $\mathsf{State}(x, y)$ states that, in time $x$, the state of $M$ is $y$, and $\mathsf{Head}(x, y)$ states that, in time $x$, the head of $M$ is under the $y$-th tape cell. The last rule says that every tape cell after the input string are initially written as "$\diamond$".

The rules that define the state transitions of $M$ are similar to those in the proof of Theorem 7.2 in (Dantsin et al. 2001). Hence, it remains to define rules for output. For each relation symbol $\mathsf{Q}_i \in \mathcal{Q}$, we define the following rule

$$\mathsf{QPos}_i(\boldsymbol{v}, x) \land \mathsf{Tape}(y, x, 1) \land \mathsf{Max}^*(y) \to \mathsf{Q}_i(\boldsymbol{v})$$

Informally, the function of these rules is to reconstruct the query (general) database from the output of $M$. Here, we assume that $M$ do the operation "nop" repeatedly so that it will terminate at the $|\mathsf{dom}(D)| + |\mathsf{dom}(D)|^k$-th stage. $\square$

**Proof of Theorem 10**

**Theorem 10.** *For all $k > 0$, there exists a $k$-exponentially bounded rule ontology $O = (\Sigma, \mathcal{D}, \mathcal{Q})$ such that, for any $(k-1)$-exponentially bounded rule ontology $O_0 = (\Sigma_0, \mathcal{D}, \mathcal{Q})$ where $\Sigma_0$ is of polynomial size w.r.t. $\Sigma$, we have $[\![O_0]\!] \not\approx [\![O]\!]$.*

*Proof.* Let $n, \ell$ and $\Sigma_{\mathsf{num}}$ be defined as in the proof of Proposition 4. Let $\mathcal{D} = \emptyset$ and $\mathcal{Q} = \{\mathsf{Min}_\ell, \mathsf{Max}_\ell, \mathsf{Succ}_\ell\}$, and let $m = \exp_{k+2}(n) - 1$. Let $Q$ be a general database over $\mathcal{Q}$ that consists of $\mathsf{Min}_\ell(\mathsf{n}_0)$, $\mathsf{Max}_\ell(\mathsf{n}_m)$ and $\mathsf{Succ}_\ell(\mathsf{n}_i, \mathsf{n}_{i+1})$ for all integers $i$ with $0 \leq i < m$, where $(\mathsf{n}_i)_{0 \leq i \leq m}$ are distinct nulls. By the analysis in the proof of Proposition 4, it is clear that $[\![(\Sigma_{\mathsf{num}}, \mathcal{D}, \mathcal{Q})]\!](\emptyset)$ is homomorphically equivalent to $Q$. (Note that $\emptyset$ is the only database over schema $\mathcal{D}$.) Since the core of $Q$ is $Q$, $Q$ should be the least universal model of $\emptyset \cup \Sigma_{\mathsf{num}}$. Towards a contradiction, we assume that there exists a $(k-1)$-exponentially bounded rule ontology $O = (\Sigma, \mathcal{D}, \mathcal{Q})$ such that $[\![O]\!] = [\![(\Sigma_{\mathsf{num}}, \mathcal{D}, \mathcal{Q})]\!]$ and $\Sigma$ is of polynomial size w.r.t. $\Sigma_{\mathsf{num}}$. Then $[\![O]\!]$ is homomorphically equivalent to $Q$. This implies that $[\![O]\!](\emptyset)$ contains at least $\exp_{k+2}(n)$ nulls. On the other hand, as $\Sigma$ is of polynomial size w.r.t. $\Sigma_{\mathsf{num}}$, by the analysis in the proof of Proposition 3, we infer that the size of $\mathsf{chase}(\emptyset, \Sigma)$ is $\exp_{k+1}(\mathcal{O}(n))$. This means that $\mathsf{chase}(\emptyset, \Sigma)$, or equivalently $[\![O]\!](\emptyset)$, contains $\exp_{k+1}(\mathcal{O}(n))$ nulls, a contradiction as desired. $\square$